\def\year{2020}\relax
\documentclass[letterpaper]{article} % DO NOT CHANGE THIS
\usepackage{aaai20}  % DO NOT CHANGE THIS
\usepackage{times}  % DO NOT CHANGE THIS
\usepackage{helvet} % DO NOT CHANGE THIS
\usepackage{courier}  % DO NOT CHANGE THIS
\usepackage[hyphens]{url}  % DO NOT CHANGE THIS
\usepackage{graphicx} % DO NOT CHANGE THIS
\usepackage{graphicx}  % DO NOT CHANGE THIS
\usepackage{boxedminipage}
\usepackage{algorithm}
\usepackage{algorithmic}
\usepackage{color}
\usepackage{amsmath}
\usepackage{graphicx}
\usepackage{subfigure}
\usepackage{multirow} 
\usepackage{makecell}
\newcommand{\ralf}[1]{\color{black} #1 \color{black}}
\newcommand{\ignore}[1]{{}}
\title{Repositioning Bikes with Carrier Vehicles and Bike Trailers in Bike Sharing Systems}
\author{Xinghua Zheng$^1$, Ming Tang$^1$, Hankz Hankui Zhuo$^1$\thanks{corresponding author}, Kevin X. Wen$^2$\\
$^1$School of Data and Computer Science, Sun Yat-Sen University, Guangzhou, China 510006\\
$^2$Department of Computer Science, University of California, Santa Cruz, California, US \\
\{zhengxh5@mail3,tangm28@mail2,zhuohank@mail\}.sysu.edu.cn; kevinxwen@gmail.com
}
 
\begin{document}

\maketitle

\begin{abstract}
Bike Sharing Systems (BSSs) have been adopted in many major cities of the world due to traffic congestion and carbon emissions. Although there have been approaches to exploiting either bike trailers via crowdsourcing or carrier vehicles to reposition bikes in the ``right'' stations in the ``right'' time, they do not jointly consider the usage of both bike trailers and carrier vehicles. In this paper, we aim to take advantage of both bike trailers and carrier vehicles to reduce the loss of demand with regard to the crowdsourcing of bike trailers and the fuel cost of carrier vehicles. In the experiment, we exhibit that our approach outperforms baselines in several datasets from bike sharing companies.
\end{abstract}

\section{Introduction}
Bike sharing systems (BSSs) typically have a set of base stations that are strategically placed throughout a city and each station has a fixed number of docks, e.g., Capital Bikeshare\footnote{https://www.capitalbikeshare.com}, Bluebikes\footnote{https://www.bluebikes.com}, Mobike\footnote{https://mobike.com/global/}, BIXI\footnote{https://montreal.bixi.com}, etc. At the beginning of the day, each station is stocked with a pre-determined number of bikes. Customers can pick and drop bikes from any station and are charged depending on the hiring duration \cite{DBLP:journals/fgcs/TsaiCH19,DBLP:conf/kdd/HulotAJ18,DBLP:conf/aips/LowalekarVGJJ17,DBLP:journals/ior/VulcanoRR12,DBLP:journals/eor/SchuijbroekHH17}. \ignore{BSSs are capable of providing healthier living and greener environments while delivering fast movements for customers ,  }
%DBLP:conf/aips/GhoshV17,

Due to the individualistic and uncoordinated movements of customers, there is often starvation (empty base stations precluding bike pickup) or congestion (full base stations precluding bike return) of bikes at certain stations, which results in a significant loss of customer demand \cite{DBLP:journals/ior/ShuCLTW13,DBLP:journals/itsm/ChenLL18}. To address this problem, a variety of systems \cite{DBLP:journals/jair/GhoshVAJ17,DBLP:conf/aips/LowalekarVGJJ17} employ the idea of repositioning idle bikes with the help of carrier vehicles during the day, by taking into account the movement of bikes by customers \cite{DBLP:journals/fgcs/TsaiCH19,DBLP:journals/tits/PfrommerWSM14,DBLP:conf/aips/GhoshV17}. 
While previous approaches of repositioning can help reduce imbalance, repositioning idle bikes using carrier vehicles (c.f. \cite{DBLP:conf/ijcai/GhoshTV16}) incurs substantial routing and fuel costs while covering entire stations\footnote{A carrier vehicle is a truck to reposition idle bikes during the day using myopic and adhoc methods so as to return to a pre-determined configuration.(e.g., each carrier vehicle can hold 30-40 bikes, its working distance is 5 kilometers away).}. In addition, repositioning idle bikes using bike trailers just carries a few of bikes once and the moving distance is limited\footnote{A bike trailer is an add-on to a bike that can carry a small number of bikes (e.g., each bike trailer can hold 3-5 bikes, its working distance is within 5 kilometers) and is useful to relocate bikes to nearby stations.}, \ignore{Meanwhile, they did not consider the available budget (c.f. \cite{DBLP:journals/jair/GhoshVAJ17}), working distance\footnote{Since nearby stations can be covered by bike trailers, we exploit the geographical proximity based clustering method to obtain main stations to reduce the usage of carrier vehicles.} and carrying capacity, }which restrict the usage of bike trailers to reposition bikes among stations.
%\footnote{Each bike trailer can hold 3-5 bikes, and each carrier vehicle can hold 30-40 bikes}
%Working distance of carrier vehicles is typically within 5 kilometers of each other for main stations respectively, and working distance of bike trailers is typically within each main station.

In this paper, we propose an optimization model called ({\tt DRRPVT}), which stands for \textbf{D}ynamically \textbf{R}epositioning and \textbf{R}outing \textbf{P}roblem with carrier \textbf{V}ehicles and bike \textbf{T}railers, to jointly consider the usage of carrier vehicles and bike trailers. We aim to better optimize the overall profit of hired bikes and consequently reduce the expected loss of demand. Specifically, we build a profit objective function to calculate the value of carrier vehicle routing (i.e., fuel cost) and bike trailers (i.e., payment for the users of bike trailers), by considering a variety of constraints with respect to carrier vehicle routing and bike repositioning. Jointly considering both carrier vehicles and bike trailers is challenging in the sense that we need to introduce new constraints to encode relations between carrier vehicles and bike trailers, and build a novel objective function to minimize the cost of repositioning (and routing) and the loss of demand. Besides, to improve the efficiency of our approach with respect to large-scale stations (as well as carrier vehicles and bike trailers), we need to design an effective mechanism for computing main base stations to help reduce the computation time. 

\ralf{In summary, our contributions are two folds.
We first propose an optimization model to improve the performance of dynamic bike repositioning by exploiting both carrier vehicles and bike trailers simultaneously, which is different from previous approaches which only consider either trailers or carrier vehicles, but not both. To do this, we build a novel profit objective function and new constraints considering relationships between carrier vehicles and bike trailers. \ignore{ to reduce lost customer demand and increase overall profits. While profit objective function has been considered for reduce lost customer demand and increase overall profits separately, previous work has not been considered jointly. }
Second, we design a clustering mechanism for computing main base stations to help improve the efficiency of solving the optimization model regarding large-scale stations and carrier vehicles and bike trailers. 
}
\ignore{
In the remainder of the paper, we first review previous work related to our work, and then address the definition of our problem. After that, we introduce all the constraints in our problem and specify the detail of our {\tt DRRPVT} approach. Finally, we present the experimental results with comparison to baselines and conclude the paper with future work.
\vspace{-0.01\textwidth}
}
\section{Related Work}
There have been many approaches proposed to deal with bike sharing issues, which can be categorized into three aspects \cite{DBLP:journals/candie/LinYC13,DBLP:conf/aips/LowalekarVGJJ17}, i.e., static repositioning using carrier vehicles, dynamic repositioning using carrier vehicles, and dynamic repositioning using bike trailers.
%DBLP:journals/ior/VulcanoRR12,
\vspace{-1.3em}
\paragraph{Static repositioning using carrier vehicles}
Static repositioning is the problem of finding routes for a fleet of vehicles to reposition bikes at the end of the day when the movements of bikes by customers are negligible, to achieve a pre-determined inventory level at the stations\cite{DBLP:journals/disopt/ChemlaMC13}. 
%Gaspero et al. provide constraint programming to efficiently solve static repositioning using large neighborhood search \cite{DBLP:journals/constraints/GasperoRU16,DBLP:journals/transci/NairM11,}. 
As user demands change frequently during the day, those approaches are not capable of dynamically adjusting the station inventory level with respect to user demands.
\vspace{-1.3em}
\paragraph{Dynamic repositioning using carrier vehicles}
To consider dynamic repositioning using carrier vehicles with respect to the movements of customers during the day, Lowalekar et al. provide a scalable online repositioning solution using multistage stochastic optimization with online anticipatory algorithms \cite{DBLP:conf/aips/LowalekarVGJJ17,DBLP:conf/iccS/WangZMH18}. Pierre et al. develop a efficient mechanism to maximize the decision intervals between repositioning events by online rebalancing operations\cite{DBLP:conf/kdd/HulotAJ18,DBLP:journals/itsm/ChenLL18}.As dynamic repositioning using vehicles alone incurs substantial routing and fuel cost, those approaches should be improved by considering self-sustaining and environment friendly. %\cite{DBLP:journals/transci/NairM11,DBLP:journals/ior/ShuCLTW13} provide dual-bounded joint-chance constraints to efficiently solve dynamic repositioning by predicting demand with certain probability.%\cite{DBLP:journals/disopt/ChemlaMC13,DBLP:journals/ior/BattarraEV14} propose branch and cut algorithm to solve the static repositioning problem.,DBLP:journals/eor/SchuijbroekHH17
\vspace{-1.3em}
\paragraph{Dynamic repositioning using bike trailers}
To consider the self-sustaining and environment issues, instead of using vehicles, Ghosh et al. propose a pricing mechanism that takes the global view of the repositioning requirements and incentives the execution of bike-trailer tasks (based on crowdsourcing) within the budget constraints \cite{DBLP:conf/aips/GhoshV17,DBLP:conf/aaai/SinglaSBMMK15}. Despite the success of those approaches, bike trailers can only take a few bikes at once and the distance of movements is limited. Besides, the value of crowdsourcing tasks may be high (over the available budget). 
%\cite{DBLP:conf/aaai/SinglaSBMMK15,DBLP:journals/tits/PfrommerWSM14} present pricing mechanisms to stimulate bike trailers for repositioning by matching each trailer to suitable stations. Furthermore,

Different from previous approaches, our {\tt DRRPVT} approach aims to leverage the advantage of using both carrier vehicles, which is able to take a large number of bikes and move to longer distance, and bike trailers, which is able to move to short distance with limited cost and allow self-sustaining, by considering the expected profit and the loss demands reduction of repositioning and routing solution \ralf{\cite{DBLP:conf/atal/HartuvAK18,DBLP:conf/rss/ZhangP14}. }
 
\section{Problem Formulation}
Our bike sharing problem is formally defined by the following tuple: $\langle \mathcal{S}, \mathcal{V}, \mathcal{F}, \mathcal{C}^{\#}, \mathcal{C}^{*}, d^{\#}, d^{*},\sigma, R,P,\hat{P},D,B \rangle$, where 
\begin{itemize}
\item $\mathcal{S}$ denotes the set of base stations.
\vspace{-0.01\textwidth}
\ralf{\item $\mathcal{V}$ denotes the set of vehicles used for repositioning which restricted to carrier vehicles only.}
\vspace{-0.01\textwidth}
\item $\mathcal{F}$ denotes samples of customer requests for the future time steps with $F_{s,s'}^{t}$ indicating the number of customer requests between stations $s$ and $s'$ which start at decision epoch $t$ and end at decision epoch $t+1$.
\vspace{-0.01\textwidth}
\item $\mathcal{C}^{\#}$ denotes the capacity of stations with $C_s^{\#}$ indicating capacity of station $s$.
\vspace{-0.01\textwidth}
\item $\mathcal{C}^{*}$ denotes the capacity of carrier vehicles with $C_v^{*}$ indicating capacity of vehicle $v$.
\vspace{-0.01\textwidth}
\item $d^{\#}$ denotes the distribution of bikes at stations with $d_s^{\#,t}$ indicating the number of bikes at station $s$ at decision epoch $t$.
\vspace{-0.01\textwidth}
\item $d^{*}$ denotes the distribution of bikes in vehicles with $d_v^{*}$ indicating the number of bikes in vehicle $v$ at decision epoch $t$.
\vspace{-0.01\textwidth}
\item $\sigma$ denotes the distribution of carrier vehicles at stations, with $\sigma_{v,s}^t$ set to be 1 if vehicle $v$ is present at station $s$ at decision epoch $t$ and 0 otherwise.
\vspace{-0.01\textwidth}
\item $R$ denotes the revenue of bikes being hired, with $R_{s,s'}^{t}$ indicating the revenue from station $s$ to $s'$ which starts at decision epoch $t$ and ends at decision epoch $t+1$.
\vspace{-0.01\textwidth}
\item $D$ denotes the actual distance with $D_{s,s'}$ indicating the distance between stations $s$ and $s'$.
\vspace{-0.01\textwidth}
\item $B$ denotes the total budget for all trailers to bid. In other words, the total amount of value spent on trailers should not be larger than $B$.
\vspace{-0.01\textwidth}
\item $\hat{P}$ denotes the value for executing the task of bike trailer with $\hat{P}_{s,{s}'}$ indicating the value for executing the task of bike trailer picking up idle bikes at station $s$ and dropping off them at station ${s}'$.
\vspace{-0.01\textwidth}
\item $P$ denotes the routing value (e.g., fuel cost) for vehicles travelling with $P_{s,s'}$ indicating the routing value for vehicles travelling from station $s$ to $s'$ which depends on the distance between the two stations.
\vspace{-0.01\textwidth}
\end{itemize}

We make the following assumptions for the ease of explanation and representation: 
\begin{enumerate}
\item We assume that users who carry bikes and trailers at decision epoch $t$ always return their bikes at the beginning of the decision epoch $t+1$. The duration of each decision epoch is 30 minutes \ralf{\footnote{We evaluate shorter duration impacts on runtime performance. Reducing the duration of time step notably increases the runtime. There is a trade-off between utility and runtime in deciding the duration of time step. Although the performance in terms of profit and lost demand decreases by a small amount for 30 minutes of time step (over 15 minutes of time step), it provides a significant computational gain and is particularly helpful when solving large problems. Therefore, we choose 30 minute as the default setting for the duration of time step.};}
\item We sampled the empirical distribution of the real historical data of customer requests to simulate customer requests for the future time steps \cite{DBLP:journals/tits/PfrommerWSM14}. We assume that the lost demand at the time of return. Once the distribution of bikes across the stations for time step $t+1$ is obtained, we utilize this information to compute the repositioning strategy for trailers and vehicles for time step $t+1$. This iterative process continues until we reach the last decision epoch;
\vspace{-0.01\textwidth}
\item \ralf{Customers can rent a bike for 30 minutes or more, and they have to know in advance at which station they will return the bike. On the other hand, they return their bikes to the nearest available station if the destination station is full, and they leave the system if they encounter an empty station.}
\end{enumerate}
The goal of our {\tt DRRPVT} approach is to maximize the expected profit over the entire time horizon. Let $U$ denotes the sum of revenue of hired bikes and the fuel cost of vehicles and the value of bike trailers. We provide an optimisation model for a given {\tt DRRPVT}. Specifically, we provide a mixed integer linear programming (MILP for short) that computes a profit maximising repositioning and routing solution. The objective is shown in Equation (\ref{objective}):
\vspace{-0.03\textwidth}
%\centerline{\large{Table 1: MILP Formulation}}% \\
\begin{center}
%\vspace{-10pt}
\begin{boxedminipage}{8.5cm}
%\vspace{-8pt}
\begin{eqnarray}\label{objective}
\max_{y,z,a,b}U= \max_{y,z,a,b}\sum_{s,{s}',t}R_{s,{s}'}^{t}\times x_{s,{s}'}^{t}- \nonumber \\
 \sum_{t,v,s,{s}'}P_{s,{s}'}\times z_{s,{s}',v}^{t}-\sum_{t,v,s,{s}'}b_{s,{s}',v}^{t}\times\hat{P}_{s,{s}'}
\end{eqnarray} 
%\begin{eqnarray}\label{objective}
%\max_{y,z,a,b}\sum_{s,{s}',t}U_s^{t}= \sum_{s,{s}',t}R_{s,{s}'}^{t}\times x_{s,{s}'}^{t}- \nonumber \\
%\sum_{t,v,s,{s}'}P_{s,{s}'}\times z_{s,{s}',v}^{t}-\sum_{t,v,s,{s}'}b_{s,{s}',v}^{t}\times\hat{P}_{s,{s}'}
%\end{eqnarray}
\quad $s.t.$  $constraints$ $C1$-$C15$  $which$ $depends$ $on$ $y,z,a,b$.
%\vspace{-8pt}
\end{boxedminipage}
\end{center}
\textbf{Objective:} To represent the trade-off between lost demand (or alternatively the revenue from customer trips) and the value $P$ of using carrier vehicles and the value $\hat{P}$ of bike trailers, we employ the dollar value of both quantities and combine them into the overall profit at any decision epoch in Equation (1).
The \textbf{notations} used in the formulation are shown:
\begin{itemize}
\item $y_{s,v}^{+,t}$ denotes the number of bikes picked up from station $s$ by vehicle $v$ at decision epoch $t$.
%\vspace{-0.01\textwidth}
\item $y_{s,v}^{-,t}$ denotes the number of bikes dropped at station $s$ by vehicle $v$ at decision epoch $t$.
%\vspace{-0.01\textwidth}
\item $z_{s,{s}',v}^{t}$ denotes whether vehicle $v$ picks up bikes from station $s$ at decision epoch $t$ and drops off at station ${s}'$ at decision epoch $t+1$.
%\vspace{-0.01\textwidth}
%present at station $s$ at decision epoch $t$.
\item $a_{s,v}^{+,t}$ denotes the number of bikes picked up from station $s$ by bike trailer $v$ at decision epoch $t$.
%\vspace{-0.01\textwidth}
\item $a_{s,v}^{-,t}$ denotes the number of bikes dropped off at station $s$ by bike trailer $v$ at decision epoch $t$.
%\vspace{-0.01\textwidth}
\item $b_{s,{s}'v}^{t}$ denotes a binary decision variable which is set to be 1 if bike trailer $v$ picks up bikes from station $s$ in at decision epoch $t$ and returns bikes to station ${s}'$ in at decision epoch $t+1$ else 0 otherwise.
%\vspace{-0.01\textwidth}
\item $x_{s,{s}'}^{t}$ denotes the number of hired bikes moving from station $s$ at decision epoch $t$ to station ${s}'$ at decision epoch $t+1$.
\end{itemize}

\section{Constraints}
In this section, we address the constraints (\textbf{C1-C15}) we exploit in our bike sharing system, where constraints(\textbf{C1-C4}) are newly created in this paper, while constraints (\textbf{C5-C8}) have presented by \cite{DBLP:conf/aips/LowalekarVGJJ17,DBLP:journals/jair/GhoshVAJ17} and constraints (\textbf{C9-C15}) have presented by \cite{DBLP:conf/aips/GhoshV17}.
\vspace{-0.035\textwidth}
\paragraph{C1: Preservation of Bike Flows in and out of station.} We require that the bike flows in and out of stations should ensure that the number of bikes $d_{s}^{\#,t+1}$ is equivalent to the sum of bikes $d_{s}^{\#,t}$ in the previous time step and the \emph{net number} of bikes coming into the station during that time step, i.e., for each station $s$ and epoch $t$,
%\begin{align}
$d_{s}^{\#,t+1}=d_{s}^{\#,t}+\sum_{\tilde{s}}x_{\tilde{s},s}^{t-1}-\sum_{{s}'}x_{s,{s}'}^{t} + \sum_{v}(y_{s,v}^{-,t} -y_{s,v}^{+,t}+a_{s,v}^{-,t} -a_{s,v}^{+,t})$ where the \emph{net number} is defined by the last three components.
\vspace{-0.025\textwidth}
\paragraph{C2: Preservation of Bikes Flows between any two stations follow the transition dynamics observed in the data.} As a subset of arrival demand can be served if the number of bikes present in a station is less than the arrival demand, we require that bikes flows between station $s$ and ${s}'$ should be less than the product of the number of bikes present in the source station $s$ ($d_{s}^{\#,t}$) and the transition probability that a bike will move from $s$ to ${s}'$ according to expected customer demand, i.e., for each $t,s,{s}'$,
$x_{s,{s}'}^{t}\leq d_{s}^{\#,t}\times \frac{F_{s,{s}'}^{t}}{\sum_{\widetilde{s}}F_{s,\widetilde{s}}^{t}}$.
\vspace{-0.035\textwidth}
\paragraph{C3: Value of task for bike trailer.} We require a mechanism for crowdsourcing the repositioning tasks to the users of bike trailers and generating a payment method to ensure that the users bid for the tasks truthfully. The valuation of trailer $v$ task is proportional to the expected lost demand reduced by the trailer job in the training demand scenario($\xi$ represents unit value of lost demand to compute overall value), i.e.,for each $s,{s}',t$,
$\hat{P}_{s,{s}'}^{t} =\xi\times\sum_{s,{s}'}(F_{s,{s}'}^{t}-d_{s}^{\#,t})$.
\vspace{-0.045\textwidth}
\paragraph{C4: Ensuring the Budget Feasibility.} We require to incentive compatibility over all tasks without violating the fix budget $B$ feasibility. Each task of trailers $v\in V $ has a valuation for the task is denoted by $\hat{P}$. We aim to allocate the tasks in a fashion that maximizes the overall valuation of the center while the total payment is bounded by the given budget $B$, i.e., 
$\sum_{s,{s}',v}b_{s,{s}',v}^{t}\times \hat{P}_{s,{s}'}^{t}\leq B$.
\vspace{-0.035\textwidth}
\paragraph{C5: Preservation of Bikes Flows in and out of vehicles.} We require that the number of bikes in a vehicle at a time step ($d_{v}^{*,t+1}$) is equivalent to the sum of the number of bikes in the vehicle at the previous time step ($d_{v}^{*,t}$) and the net number of bikes coming into the vehicle during that time step ($\sum_{s}(y_{s,v}^{+,t}-y_{s,v}^{-,t})$, i.e., for each $v,t$,
$d_{v}^{*,t}+\sum_{s}(y_{s,v}^{+,t}-y_{s,v}^{-,t})=d_{v}^{*,t+1}$.
\vspace{-0.03\textwidth}
\paragraph{C6: Preservation of Vehicles Flows in and out of stations.} We require that the number of vehicles going out of station s ($\sum_{s’} z_{s’,s,v}^{t-1}$) plus the number of vehicles present at station s at time epoch t-1 ($\sigma_{s,v}^{t-1}$) is equivalent to the sum of the number of vehicles coming into station s ($\sum_{s’} z_{s,s’,v}^{t}$) and the vehicles which are present at station s at time epoch t ($\sigma_{s,v}^{t}$). Note that one of $\sum_{s’} z_{s,s,v}^{t}$ and $\sigma_{s,v}^{t}$ could be one at most, i.e.,for each $t,s,v$,
$\sum_{s’} z_{s,s’,v}^{t} + \sigma_{s,v}^{t} = \sum_{s’} z_{s’,s,v}^{t-1} + \sigma_{s,v}^{t-1}$.
\vspace{-0.03\textwidth}
\paragraph{C7: A maximum of one vehicle can be present in one station at any time step.} Due to limited space availability near base stations and to avoid a synchronisation issue in pickup or drop-off events by multiple vehicles from the same station at the same time step, we require that the maximum number of vehicles at a station ($\sum_{{s}',v}z_{s,{s}'v}^{t}$) less than 1, i.e.,for each $t,s$,
$\sum_{{s}', v}z_{s,{s}',v}^{t} \leq 1$.
\vspace{-0.025\textwidth}
\paragraph{C8: Vehicles can only pick up or drop off bikes at a station if they are present at that station.} We require that the number of bikes picked up or dropped off at station at each time step by each vehicle is bounded by whether the station is visited by the vehicle at that time step or not, i.e.,for each $s,v,t$,
$y_{s,v}^{+,t}+y_{s,v}^{-,t}\leq C_v^*\times \sum_{{s}'}z_{s,{s}',v}^{t}$.
\vspace{-0.025\textwidth}
\paragraph{C9: Trailer capacity is not exceeded while picking up bikes.} We require that the number of bikes picked up by trailer $v$ from station $s$ is bounded by the minimum value between the number of bikes present in the station and the capacity of the trailer. $b_{s,{s}',v}$ denotes a binary decision variable which is set to 1 if bike trailer $v$ picks up bikes from station $s$ and drop off bikes to any station ${s}'$ and 0 otherwise, i.e.,for each $s,v,t$, 
$a_{s,v}^{+,t}\leq \sum_{{s}'}b_{s,{s}',v}^{t}\times min(d_{s}^{\#,t},C_{v}^{*})$.
\vspace{-0.025\textwidth}
\paragraph{C10: Total number of bikes picked up from a station is less than the available bikes.} As multiple trailers can pick up bikes from the same station, we require that the total number of picked up bikes by all the trailers from station s during the planning period $t$ is bounded by the number of bikes present at the station ($d_{s}^{\#,t}$), i.e.,for each $s,t$,
$\sum_{v}a_{s,v}^{+,t}\leq d_{s}^{\#,t}$. 
\vspace{-0.025\textwidth}
\paragraph{C11: Station capacity is not exceeded while dropping off bikes.} We require that the total number of dropped off bikes at station $s$ is bounded by the number of available slots for bikes at that station, i.e.,for each $s,t$,
$\sum_{v}a_{s,v}^{-,t}\leq C_{s}^{\#}-d_{s}^{\#,t}$.
\vspace{-0.025\textwidth}
\paragraph{C12: Total travelling distance for a trailer is bounded by a threshold value.} To represent the physical limitation of route, we need to ensure that the total distance travelled by a trailer in a given planning period is within a few kilometers. We require that the distance between pick-up station and the drop-off station for a trailer $v$ is bounded by a threshold value, $D_{max}$, i.e.,for each $s,{s}',v,t$,
$b_{s,{s}',v}^{t}\times D_{s,{s}'}\leq D_{max}$.
\vspace{-0.04\textwidth}
\paragraph{C13: A trailer can only pick up or drop off bikes at exactly one station.} We require that a trailer can go to exactly one station starting from a specific station, i.e., for each $v,t$,
$\sum_{s,{s}'}b_{s,{s'},v}^{t}=1$.
\vspace{-0.025\textwidth}
\paragraph{C14: A trailer should return the exact number of bikes picked up.} We require that the number of bikes dropped off by a bike trailer in a station is exactly equals to the number of picked up bikes if the station is visited, i.e.,for each ${s}',v,t$, 
$a_{{s}',v}^{-,t}=\sum_{s}(b_{s,{s}',v}^{t}\times a_{s,v}^{+,t})$. 
Note that, above equation are non-linear in nature. However, one component in the right hand side is a binary variable. Therefore, we can easily linearize them using the following formula, i.e.,for each ${s}',v,t$,
$a_{{s}',v}^{-,t}\leq C_{v}^{*}\times  \sum_{s}b_{s,{s}',v}^{t}$, 
$a_{{s}',v}^{-,t}\leq \sum_{s}a_{s,v}^{+,t}$, 
$a_{{s}',v}^{-,t}\geq \sum_{s}a_{s,v}^{+,t}-( 1-\sum_{s}b_{s,{s}',v}^{t})\times C_{v}^{*}$. 
\vspace{-0.03\textwidth}
\paragraph{C15: Station and vehicle capacities are not exceeded when repositioning bikes.} We require that the number of bikes at a station $s$ does not exceed the number of available docks at that station ($C_{s}^{\#}$). Similarly, these constraints also enforce that the number of bikes picked up or dropped off by a vehicle $v$ in aggregate does not exceed the capacity of the vehicle ($C_{v}^{*}$), i.e., 
$0\leq x_{s,v}^{t}\leq F_{s,v}^{t};0\leq d_{s}^{\#,t}\leq C_{s}^{\#}; 0\leq d_{v}^{*,t}\leq C_{v}^{*}$, 
$0\leq y_{s,v}^{+,t},a_{s,v}^{+,t}\leq d_{s}^{\#,t};0\leq y_{s,v}^{-,t},a_{s,v}^{-,t}\leq C_{s}^{\#}-d_{s}^{\#,t}$, 
$0\leq y_{s,v}^{+,t},y_{s,v}^{-,t},a_{s,v}^{+,t},a_{s,v}^{-,t}\leq C_{v}^{*};z_{s,v}^{t}, b_{s,{s}',v}^{t} \in \left \{0,1 \right \}$. \\
Given \textbf{C1}-\textbf{C15}, our task is to calculate which vehicles reposition bikes from state $s$ to $s'$, i.e., $z$, and which trailers reposition bikes from $s$ to $s'$, i.e., $b$, by optimizing Equation (\ref{objective}).
%\vspace{-0.01\textwidth}
\section{Our {\tt DRRPVT} Approach}
\ralf{In order to solve Equation (1), we use the well-known Lagrangian dual decomposition (LDD) (Fisher,1985; Gordon, et al., 2012) technique. While this is a general purpose approach, its scalability, usability and utility depend significantly on the following characteristics of the model: 
%\begin{enumerate}
%\item  
\\
\textbf{Identifying the right constraints to be dualized:} This step is crucial to ensure that the resulting subproblems are easy to solve and the resulting bound derived from the dual solution is tight during the LDD process. If the right constraints are not dualized, then the underlying Lagrangian based optimization may not be decomposable or it may take significantly more time than the original MILP to find the desired solution. 
\\
%\item 
\textbf{Extraction of a primal solution from an infeasible dual solution:} The primal extraction process is important to derive a valid bound (heuristic solution) during the LDD process. In many cases, the solution obtained by solving the decomposed dual slaves can be infeasible with respect to the original formulation and hence, the overall approach can potentially lead to slower convergence and poor solutions. 
\\
%\item 
\textbf{Decompose the original problem into a master problem and two slaves
(SOLVEREDEPLOY and SOLVEROUTING):} As highlighted in Equation (1), only constraints (8) contain a dependency between routing and repositioning variables. We dualize constraints (8) using the dual variables, $\alpha _{s,t,v}$ and obtain the Lagrangian function as Equation (2).
%\end{enumerate}
}

We exploit LDD to provide a near optimal solution for the dynamic repositioning of bikes \cite{DBLP:journals/jair/GhoshVAJ17,DBLP:conf/aips/GhoshVAJ15}. \ralf{Although the LDD framework was indeed used in Ghosh et al, 2015 and 2017, challenging to investigate the usage of LDD to accommodate the new constraints.}An overview of {\tt DRRPVT} is shown in Algorithm 1. We will present main steps of Algorithm \ref{framework} in the following subsections.

%In order to solve Equation (1), we use the well-known Lagrangian dual decomposition (LDD) (Fisher,1985; Gordon, et al., 2012) technique. First, Identifying the right constraints to be dualized. Secondly, Extraction of a primal solution from an infeasible dual solution. Lastly, Decompose the original problem into a master problem and two slaves (SOLVEREDEPLOY and SOLVEROUTING). As highlighted in Equation (2), only constraints (8) contain a dependency between routing and repositioning variables. We dualize constraints (8) using the dual variables, $\alpha{s;t;v}$ and obtain the Lagrangian function as Equation (2), the pseudo code for the LDD is provided in Algorithm (1). 

%We exploit Lagrangian dual decomposition (called LDD) [Fisher,1985; Gordon, et al., 2012] to provide a near optimal solution for the dynamic repositioning of bikes \cite{DBLP:conf/aaai/KumarWZ12,DBLP:journals/jair/GhoshVAJ17}. While this is a general purpose approach, its scalability, usability and utility depend significantly on the characteristics of {\tt DRRPVT} is shown in Algorithm \ref{framework}. %We will present main steps of Algorithm \ref{framework} in the following subsections.

\begin{algorithm}
\caption{An overview of our {\tt DRRPVT} approach}
\label{framework}
\textbf{Input:} $\langle \mathcal{S}, \mathcal{V}, \mathcal{F}, \mathcal{C}^{\#}, \mathcal{C}^{*}, d^{\#}, d^{*},\sigma,P, \hat{P},R,D,B \rangle$ \\
\textbf{Output: $y, z, a, b$} 
\begin{algorithmic}[1]
\STATE $\widetilde{S}=CalculateMainStations$($\mathcal{S},D$) 
\STATE $\alpha = 0$, $iter = 0$ 
\WHILE {$[ p -( \rho1+\rho2 )]\leq \delta $}
   \STATE $\rho1,y,a,b\leftarrow SolveReposition(\alpha ^{iter})$ 
   \STATE $\rho2,z \leftarrow SolveRouting(\alpha ^{iter})$
   \STATE $\alpha^{iter+1}\leftarrow[\alpha^{iter} +\gamma \times (y^{+}+y^{-} -C^{*}\times \sum_{\widetilde{s}}z_{\cdot,\widetilde{s},\cdot})]_{+} $
   \STATE $ p,y_{p},a_{p},b_{p} \leftarrow ExtractPrimal(z)$
   \STATE $ iter\leftarrow iter+1 $
\ENDWHILE
\STATE $SolvingIncentivizeTrailerTask$($d^{\#},F,a$)
\end{algorithmic}
\end{algorithm}

Our task is to optimize Equation (\ref{objective}) to calculate $y,a,b,z$. To do this, based on Equation (\ref{objective}), we can define a Lagrangian function as shown below: 
\begin{align}
L(\alpha) = \min_{y,z,a,b}[\textrm{-}\sum_{t,s,{s}'}R_{s,{s}'}^{t}*x_{s,{s}'}^{t}\textrm{+} \sum_{t,v,s,{s}'}z_{s,{s}',v}^{t}*P_{s,{s}'}\textrm{+} \quad \quad \notag \\
\sum_{s,{s}',v,t}b_{s,{s}',v}^{t}*\hat{P}_{s,{s}'} \textrm{+} \sum_{s,v,t}\alpha _{s,t,v}( y_{s,v}^{+,t}\textrm{+}y_{s,v}^{-,t}\textrm{-}C_{v}^{*}* \sum_{{s}'}z_{s,{s}',v}^{t})] \notag 
\end{align}
\begin{align}
s.t. \ Constraints \ C1-C7 \ and \ C9-C15 \quad\quad %\nonumber
\end{align}
%\vspace{-0.075\textwidth}
%\begin{align*}
%\end{align*}
, which is equivalent to 
\begin{align}\label{lg}
L(\alpha) = \min_{y,a,b} [\textrm{-}\sum_{t,k,s,{s}'}R_{s,{s}'}^{t}* x_{s,{s}'}^{t}\textrm{+}\sum_{s,v,t}\alpha _{s,t,v}( y_{s,v}^{+,t}+y_{s,v}^{-,t} )\textrm{+} \nonumber \\
\sum_{s,{s}',v,t}b_{s,{s}',v}^{t}*\hat{P}_{s,{s}'}]\textrm{+}\min_{z}[\sum_{s,{s}',v,t}z_{s,{s}',v}^{t}*(P_{s,{s}'}\textrm{-}C_{v}^{*}*\alpha _{s,t,v})] \notag
\end{align}
%\vspace{-0.05\textwidth}
\begin{align}
s.t. \ Constraints \ C1-C7 \ and \ C9-C15 \quad\quad\quad\quad %\nonumber
\end{align}

\subsection{Calculating Main Stations}
Since nearby stations can be covered by bike trailers, we exploit the geographical proximity based clustering method to obtain main stations to reduce the usage of carrier vehicles \cite{DBLP:journals/jair/GhoshVAJ17,DBLP:journals/constraints/GasperoRU16} . \ralf{We thus provide a clustering mechanism to calculate main stations in Step 1 of Algorithm \ref{framework}. The high-level idea is to first calculate distances between base stations, and then cluster base stations based on their distances using off-the-shelf clustering approaches such as k-means. We denote the set of resulting main stations by $\widetilde{S}$ \cite{DBLP:conf/aaai/GhoshV18,DBLP:conf/aips/KondaGV18,DBLP:conf/aaai/JhaCLWRTVTR18}. Therefore, we utilize carrier vehicles to reposition bikes dynamically for a wide range (i.e., among main stations) and utilize bike trailers to reposition the bikes dynamically for a small range (i.e., within each main station).}
%DBLP:journals/advcs/BorgnatAFRRF11,DBLP:journals/transci/NairM11,DBLP:journals/ior/VulcanoRR12

\subsection{Repositioning Bikes and Routing for Vehicles}
Our goal is to design a mechanism to incentivize task execution based on the maximization of profit via dynamically repositioning and routing. Specifically, we provide a decomposition approach to exploit the minimal dependency that exists in the model DRRPVT between the repositioning problem (how many bikes to pick up and drop off at each station) and the routing problem (how to move vehicles between base stations to pick up or drop off bikes). The following observation highlights this minimal dependency: 
\begin{itemize}
\item $y,a,b$ capture the solution to the repositioning problem. 
\item $z$ captures the solution to the routing problem. 
\end{itemize}
These sets of variables only interact with each other in constraint (8). In all of the other constraints of our {\tt DRRPVT} model, the routing variables are completely independent with repositioning variables.
%DBLP:journals/ior/KokF07,

With the minimal dependency observation, we use LDD in {\tt DRRPVT}. It is crucial to ensure that the resulting subproblems are easy to solve and the resulting bound derived from the dual solution is tight during the LDD process. We first decompose the original problem into a master problem (i.e., Equation (\ref{lg})) and two slaves \emph{SolveReposition} and \emph{SolveRouting}. As highlighted, only constraint (8) contains dependencies between routing and repositioning variables, i.e., $\alpha _{v,s,t}$. Thus, we dualize constraint (8) using the dual variables, and obtain the Lagrangian function in Equation (\ref{lg}). The first three terms in Equation (\ref{lg}) corresponding to the repositioning problem are given in Equation (\ref{reposition}), and the last term corresponding to the routing problem is given in Equation (\ref{routing}), respectively, i.e.,
%\vspace{-0.03\textwidth}
\begin{center}
%\vspace{10pt}
%\begin{boxedminipage}{8.5cm}
%\vspace{-8pt}
\begin{align}\label{reposition}
\min_{y,a,b}\sum_{s,t,v}\alpha _{s,v,t}* y_{s,v}^{t}\textrm{+}\sum_{s,{s}',t,v}b_{s,{s}',v}^{t}* \hat{P}_{s,{s}'}\textrm{-}\sum_{t,s,{s}'}R_{s,{s}'}^{t}* x_{s,{s}'}^{t}  \notag 
\end{align}
\begin{align}
s.t. \ Constraints \ C1-C5 \ \ and \ C9-C15 \quad\quad %\quad\quad \nonumber
\end{align}
%\end{boxedminipage}
\end{center}
and 
%\centerline{\large{Table 3: SOLVEROUTING}} 
%\vspace{-0.02\textwidth}
%\begin{center}
%\vspace{-5pt}
%\begin{boxedminipage}{7cm}
%\vspace{-4pt}
\begin{align}\label{routing}
\min_{z}\sum_{v,s,\widetilde{s}}^{t}z_{s,\widetilde{s},v}^{t}\times \left ( P_{s,\widetilde{s}}-C_{v}^{*}\times \alpha _{s,t,v} \right ) \nonumber  \\
s.t. \quad Constraints \ C6\textrm{-}C7 \ and \ C15
\end{align}
%\end{boxedminipage}
%\end{center}
From Equation (\ref{lg}), given $\alpha$, the dual value corresponding to the original problem is obtained by adding up the objective function values from the two slaves, which yields a valid lower bound with respect to the original problem. It should be noted that the decomposition is only for $L(\alpha)$. The value of \emph{SolveReposition} is denoted by $\rho_1$, and  The value of \emph{SolveRouting} is denoted by $\rho_2$.

Next, we solve the following optimization problem at the \textbf{master} in order to reduce violations of the dualized constraints: $\max_{\alpha \geq 0}L\left ( \alpha  \right )$.
This \textbf{master} optimization problem is solved iteratively using a sub-gradient descent method applied on the dual variables $\alpha $, i.e., Step 6 of Algorithm \ref{framework}, where $\gamma $ is a step-size parameter. The algorithm terminates when the difference between the primal objective (defined as $p$ in Algorithm 1) and the dual objective (the sum of the slave’s objectives $\rho_1,\rho_2$) is less than a pre-determined threshold value $\delta $. In order to compute the best primal solution in conjunction with the dual solution, it is important to obtain a primal solution after each iteration from the solutions of the slaves. 
%\subsection{Routing for Vehicles}
%\subsection{Extract Primal}
%We aim to extract the primal solution from an infeasible dual solution.Due to the primal extraction process is important to derive a valid bound (heuristic solution) during the LDD process,the solution obtained by solving the decomposed dual slaves can be infeasible with respect to the original formulation. Hence, the overall approach can potentially lead to slower convergence and poor solutions.
The infeasibility in the dual solution arises because the routes of the vehicles (obtained by solving the routing slave) may not be consistent with the repositioning plan of bikes (obtained by solving the repositioning slave). However, the solution for the routing slave is always feasible and can be fixed to obtain a feasible primal solution with respect to the original problem. Let $z_{s,v}^{t}=\sum_{{s}'}z_{s,{s}',v}^{t}$. We extract the primal solution by solving the optimization formulation in Equation (\ref{primary}):
%\vspace{-0.03\textwidth}
%\vspace{-10pt}
%\centerline{\large{Table 4: EXTRACTPRIM}} 
\begin{center}
\begin{boxedminipage}{8cm}
%\vspace{-8pt}
\begin{align}\label{primary}
\max_{y}&\sum_{t,s,{s}'}R_{s,{s}'}^{t}\times x_{s,{s}'}^{t} -\sum_{t,v,s,{s}'}b_{s,{s}',v}^{t}\times\hat{P}_{s,{s}'}  \nonumber \\
&s.t. \ Constraints \ C1-C5 \ \ and \ C9-C15 \\
&y_{s,v}^{+,t}+y_{s,v}^{-,t}\leq C_{v}^{*}\times z_{s,v}^{t} \quad  \forall t,s ,v   \nonumber 
\end{align}
\end{boxedminipage}
\end{center}
Specifically, constraints in Equation (\ref{primary}) are equivalent to constraint (8) where we use the solution values of the routing slave $z$ as the input. Thus, \emph{ExtractPrimal} satisfies C1-C5,C9-15 and produces a feasible solution to the original problem. Finally, we subtract the routing value from the objective value to get the correct primal value
%The error in the solution quality obtained by the Lagrangian dual decomposition method in Algorithm 5 is bounded by the difference between the primal objective, $p$ and the dual objective,($\rho 1,\rho 2$ ).

\subsection{Incentivize Trailer Tasks}
In Step 10, we use an incentivizing mechanism proposed by \cite{DBLP:conf/aips/GhoshV17,DBLP:journals/sigecom/Cavallo09}, which allocates the tasks to users of bike trailers. Firstly, the mechanism computes the value of the tasks according to the lost demand reduced by the trailer task. Secondly, it employs an incentive compatible mechanism that ensures users always bid truthfully on each task . Finally, it assigns the task to a bidder so that the profit is maximized, and employs a payment method to ensure that the task is always allocated to the lowest bidder. The total payment given to the users of trailers due to the resulting allocation should respect to the given budget $B$.

\section{Experiments}
To exhibit the effectiveness of our approach, we conducted the experiment on two datasets {\sl Capital Bikeshare }\footnote{http://www.capitalbikeshare.com/system-data} and {\sl Hubway}\footnote{http://hubwaydatachallenge.org/trip-history-data/}, and a synthetic dataset which was derived from multiple real datasets. We generated the synthetic dataset by first taking a subset of the stations from the two real-world datasets, and then taking customer demand, station capacity, geographical location of stations, initial distribution, bid values and value model drawn from the two real-world datasets. %, (3) generating the bid values from the real-world datasets.
\ralf{The \emph{Hubway} dataset consists of 95 base stations and 3 vehicles, 10 trailers; \emph{Capital Bikeshare} dataset consists of 305 active stations and 10 vehicles, 35 trailers; and the synthetic dataset consists of 60 base stations, 2 vehicles, 7 trailers. 
We employed k-means clustering to generate 12 main stations (5 base stations are grouped into 1 main stations) which are within 5 kilometers between each other. We took 6 hours of planning horizon in the morning peak (5AM-12PM) and 31 hours of planning horizon in the whole day (5AM-12AM). The duration of each decision epoch was set to be 30 minutes. The demand scenarios were collected from three months of historical trip data. Once the distribution of bikes and vehicles from stations at time step $t$ is obtained, the information is utilized to compute the repositioning strategy for trailers at time step $t+1$. \ignore{This iterative process continues until we reach the last decision epoch, as shown in the second paragraph of Section Problem Formulation (i.e., assumptions 1-3).}}

Let $G_v$ and $G_t$ denote the gains of profit with {\tt DRRPV} and {\tt DRRPT}, respectively, and $L_v$ and $L_t$ denote lost demand reductions of {\tt DRRPV} and {\tt DRRPT}, respectively. We compute $G_v$ and $G_t$ as shown below: $G_v=\frac{\mathcal{U}_{vt}-\mathcal{U}_{v}}{\mathcal{U}_{v}}$, $G_t=\frac{\mathcal{U}_{vt}-\mathcal{U}_{t}}{\mathcal{U}_{t}}$, $L_v=\frac{\mathcal{E}_{vt}-\mathcal{E}_{v}}{\mathcal{E}_{v}}$, $L_t=\frac{\mathcal{E}_{vt}-\mathcal{E}_{t}}{\mathcal{E}_{t}}$, where $\mathcal{U}_{vt}$ and $\mathcal{E}_{vt}$ indicate the profit and lost demand reduction of using both carrier vehicles and bike trailers, respectively; $\mathcal{U}_{v}$ and $\mathcal{E}_{v}$ indicate the profit and lost demand reduction of using carrier vehicles only, respectively; $\mathcal{U}_{t}$ and $\mathcal{E}_{t}$ indicate the profit and lost demand reduction of using bike trailers only, respectively.

\ignore{
In terms of scalability, LDD improves over MILP, and the use of main stations in LDD further improves the  performance of LDD. The runtime is primarily employed to measure scalability and whether we are able to get a high quality solution within a reasonable amount of time. The duality gap is significantly employed to measure scalability and whether we are able to get the convergence of LDD to near optimal solutions. }

We would like to verify the following aspects\footnote{All optimization models were solved based on GUROBI 7.5.2 and 4.0 GHz Intel Core i7.}. We first evaluate that our {\tt DRRPVT} approach with novel mechanism (LDD + Main station) outperforms two baselines which use vehicles \cite{DBLP:conf/aips/LowalekarVGJJ17} and trailers \cite{DBLP:conf/aips/GhoshV17}, respectively. We then compare LDD and main stations  in {\tt DRRPVT} with MILP to see the advantage of LDD and Main stations. We finally evaluate {\tt DRRPVT} remains robust with respect to variation of the numbers of stations, vehicles and trailers.

\subsection{Experimental Results}
\subsubsection{Comparison against Baselines}
We provide the key performance comparison with respect to the overall profit to show that we can reduce the lost demand without incurring extra value to the operators. We employ 3 vehicles and 20 bike trailers for the experiments in both {\sl Capital Bikeshare} and  {\sl Hubway}, which is also exploited by \cite{DBLP:journals/jair/GhoshVAJ17}. We evaluate {\tt DRRPVT} with respect to different time periods, i.e., the peak period and the whole day. 
\begin{table}
\caption{Comparison of profit gain and lost demand reduction in a whole day (5AM-12AM)} \label{whole-day}
\centering  
\small
\begin{tabular}{|c|c|c|c|c|}
\hline
datasets & $\mathcal{G}_{v}$ & $\mathcal{L}_{v}$ &$\mathcal{G}_{t}$ & $\mathcal{L}_{t}$ \\
\hline 
Hubway & 2.42\% & 23.57\% & 2.18\% & 26.91\%  \\
\hline 
Capital Bikeshare & 1.97\% & 14.42\% & 1.25\% & 17.38\%  \\
\hline
\end{tabular}
\end{table}
\begin{table}
\caption{Comparison of profit gain and lost demand reduction in the peak period (5AM-12PM)} \label{peak}
\centering  
\small
\begin{tabular}{|c|c|c|c|c|}
\hline
datasets &  $\mathcal{G}_{v}$ & $\mathcal{L}_{v}$ &$\mathcal{G}_{t}$ & $\mathcal{L}_{t}$ \\
\hline 
Hubway & 4.63\% & 29.71\% & 4.26\% & 31.12\%  \\
\hline 
Capital Bikeshare & 4.25\% & 19.39\% & 4.11\% & 24.45\%  \\
\hline
\end{tabular}
\end{table}
%\vspace{-0.01\textwidth}
\ralf{Tables 1 and 2 show the average percentage gain in profit and reduction in lost demand with our approach in comparison to the baselines on the two real-world datasets. Based on the aggregate results, our approach {\tt DRRPVT} is always able to outperform both {\tt DRRPV} and {\tt DRRPT} with respect to both of the profit gain and lost demand reduction. From Table 1, our approach performs much better in {\sl Hubway} than  {\sl Capital Bikeshare} comparing to baselines. This is because the number of users hiring bikes in {\sl Hubway} is much larger than  {\sl Capital Bikeshare}. The more users hire bikes, the better our approach performs. Similar results can be found in Table 2.}
%(short for \textbf{D}ynamically \textbf{R}epositioning and \textbf{R}outing \textbf{P}roblem with carrier \textbf{V}ehicles) and {\tt DRRPT}(short for \textbf{D}ynamically \textbf{R}epositioning and \textbf{R}outing \textbf{P}roblem with bike \textbf{T}railers)
\begin{figure*}[bhtp]   
  \begin{minipage}[b]{0.33\textwidth} % 如果一行放2个\UTF{56FE}，用0.5，如果3个\UTF{56FE}，用0.33  
    \centering   
    \includegraphics[width=1.0\textwidth]{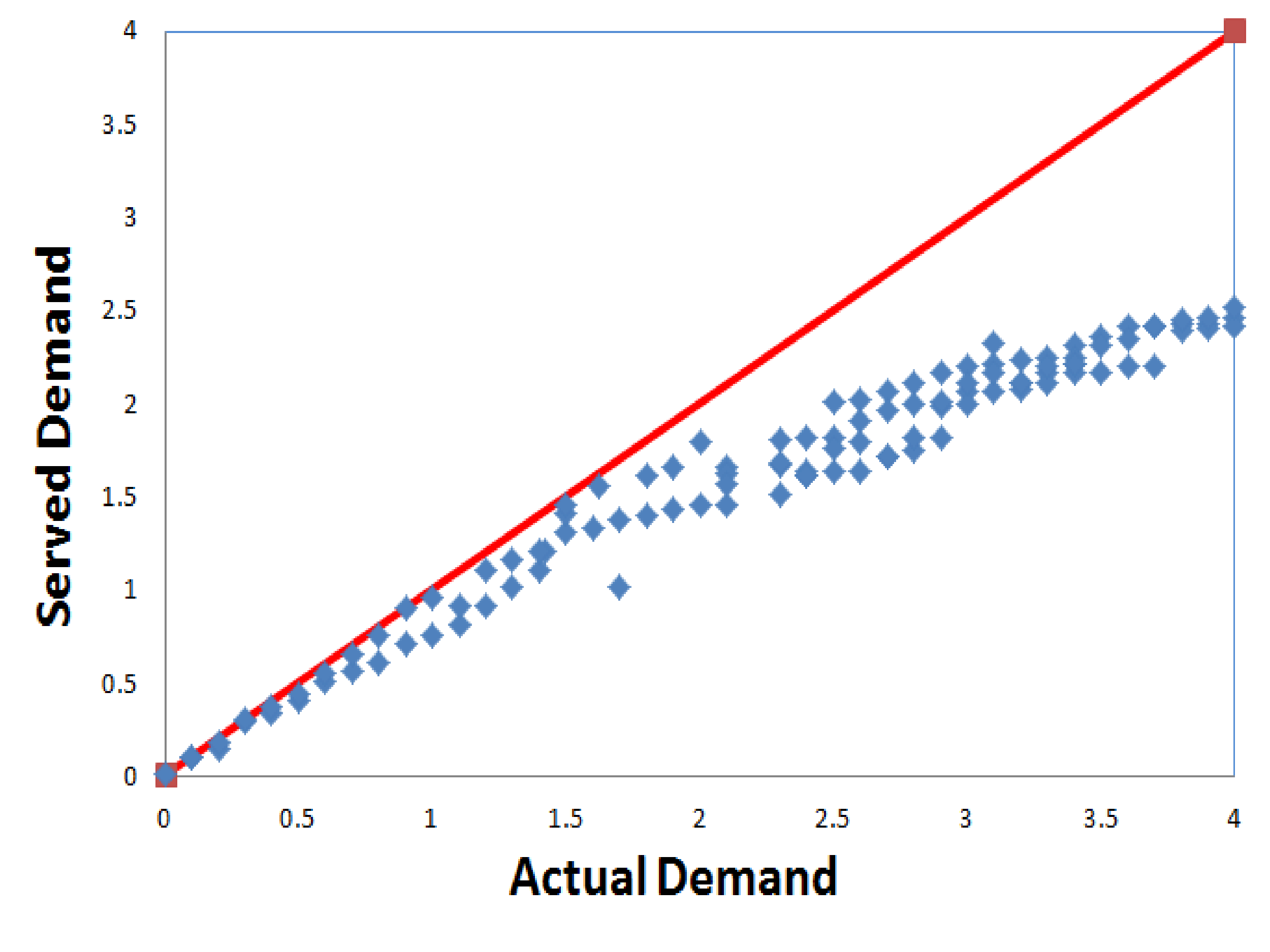}  
      \centerline{(a)}    
  \end{minipage}%  
  \hfill  %水平填充
  \begin{minipage}[b]{0.33\textwidth} % 如果一行放2个\UTF{56FE}，用0.5，如果3个\UTF{56FE}，用0.33  
    \centering   
    \includegraphics[width=1.0\textwidth]{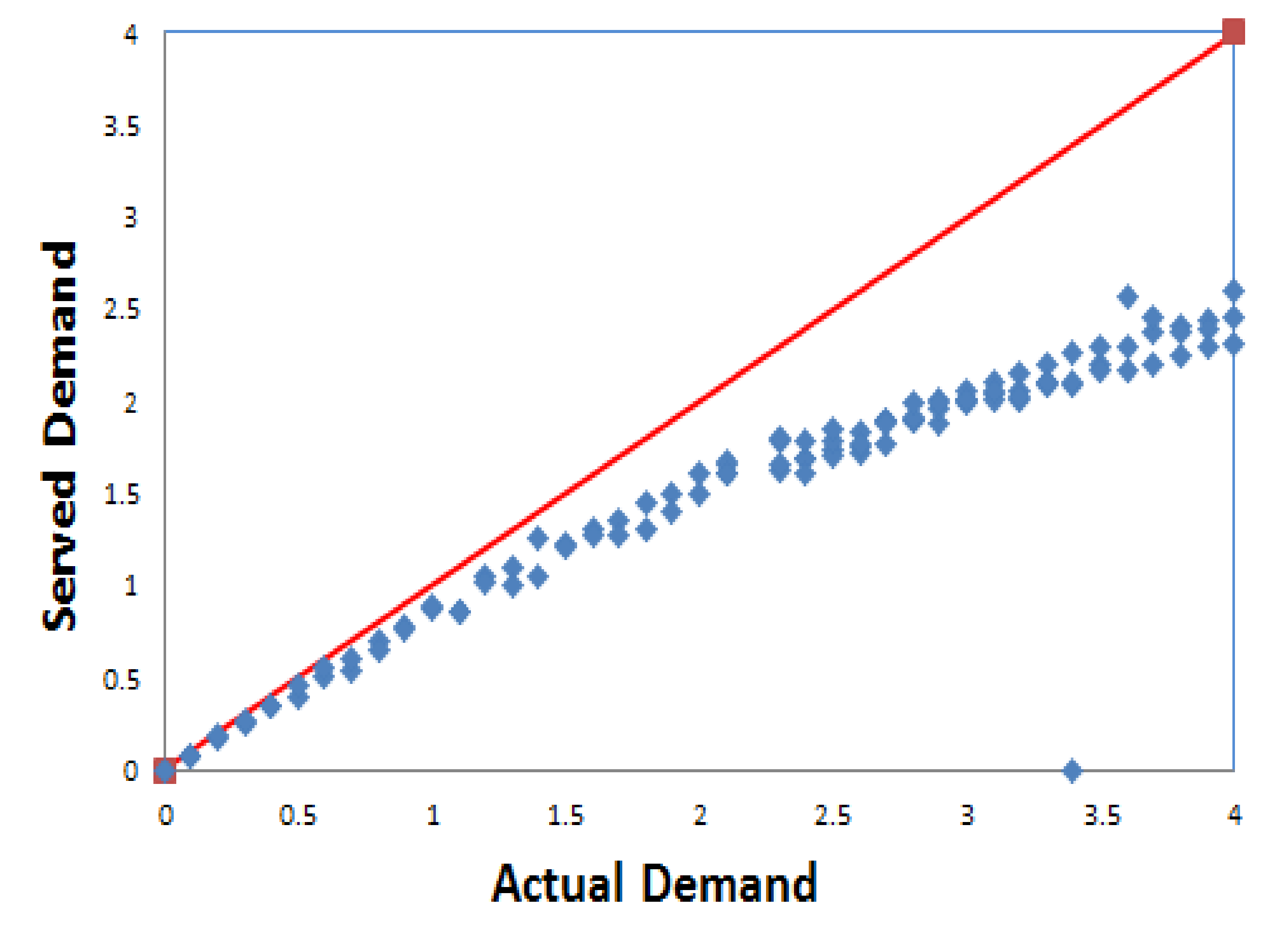}  
      \centerline{(b)}    
  \end{minipage}%  
  \hfill  %水平填充
  \begin{minipage}[b]{0.33\textwidth} % 如果一行放2个\UTF{56FE}，用0.5，如果3个\UTF{56FE}，用0.33   
    \centering   
    \includegraphics[width=1.0\textwidth]{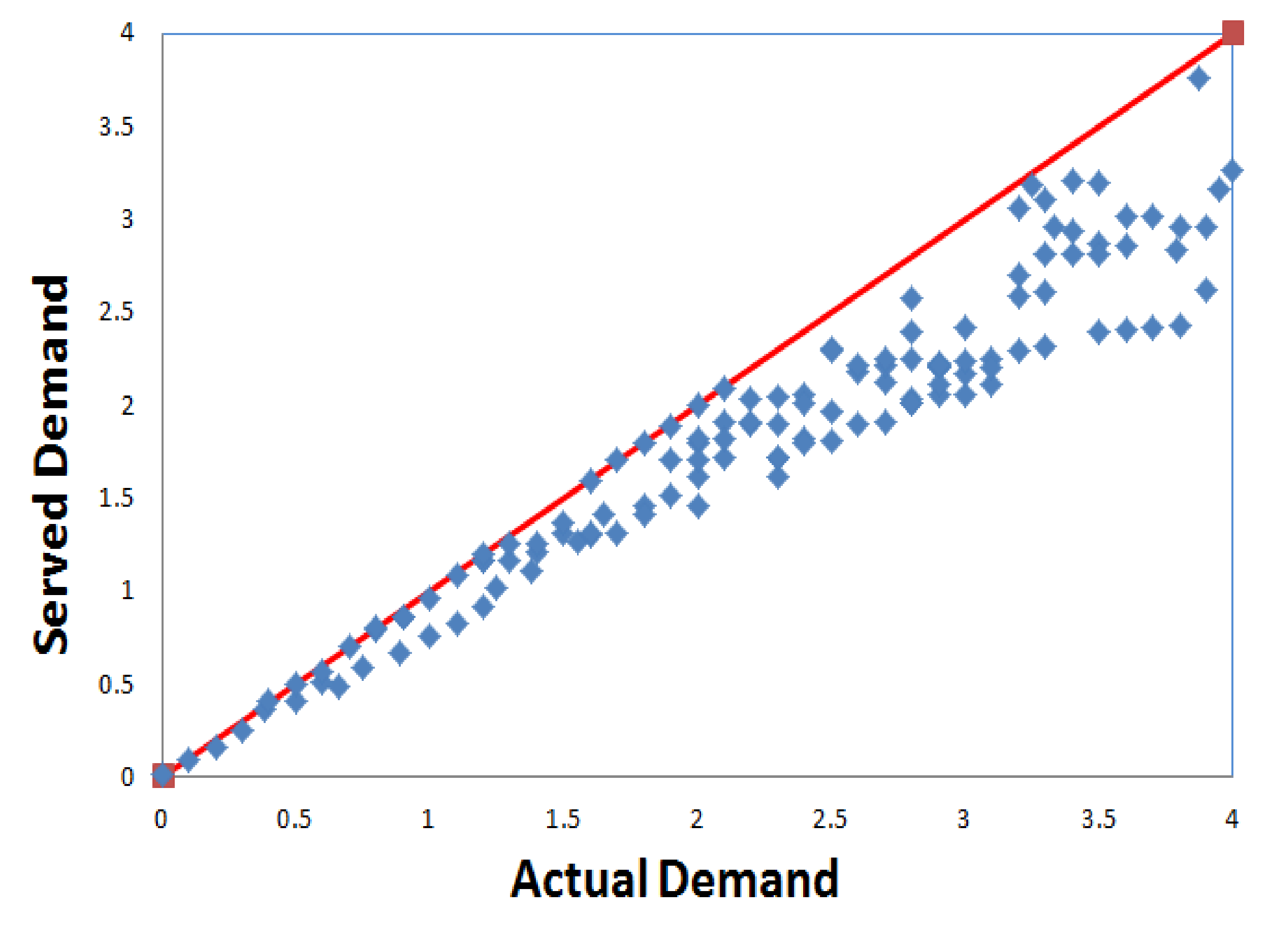} 
      \centerline{(c)}      
  \end{minipage}
  \hfill  %水平填充
  \caption{Correlation of demand and supply: (a) DRRPV, (b) DRRPT, and (c) DRRPVT} \label{fig:1}
\end{figure*}
%\vspace{-0.02\textwidth}

Lastly, to see the effect of repositioning, we draw the correlation between the actual demand and the served demand over decision epoch. \ralf{Figure (\ref{fig:1}) shows the correlation by running the three approaches. Each point in the figure corresponds to the values of an actual demand and its corresponding served demand for all time steps and all stations in the {\sl Hubway} data set. As expected, our approach has significantly more points closer to the identity line than the other two, which indicates our approach is able to better match the supply of bikes with the demand for bikes.} %Figure (1) shows the correlation by running the three approaches. Each point in the figure corresponds to the values of an actual demand and its corresponding served demand for all time steps and all stations in the {\sl Hubway} data set. As expected, our approach has significantly more points closer to the identity line than the other two, which indicates our approach is able to better match the supply of bikes with the demand for bikes. Therefore, it is better if more points are closer to the identity line $(x = y)$. Our experimental results demonstrate that our approach is highly competitive in terms of gain of the overall profit and lost demand reduction.

\ralf{
\subsubsection{Comparison with MILP}
We next compare LDD and Main stations of {\tt DRRPVT} to MILP with respect to runtime performance, duality gap and main stations. \ignore{Based on the same budgets and resources, LDD can solve larger scale problems than MILP.} %We used for separate experiments using MILP and LDD. 
\\
\textbf{Runtime performance:} We compare the runtime of {\tt DRRPVT}  with MILP, as shown in Figure (\ref{fig:2}a). The X-axis denotes the number of stations from 5 to 60. The Y-axis denotes the total time taken to solve problem in seconds. We can see that {\tt DRRPVT} generally outperforms MILP with respect to number of stations. MILP is unable to finish within a cut-off time of 3 hours for any problem with more than 20 stations, while {\tt DRRPVT} is able to obtain near optimal solutions on problems with 60 stations in less than 3 hour. \ralf{{\tt DRRPVT} becomes relatively stable after reaching 35 stations (the red curve). It could be easily speeded up by running our approach in a server of higher performance in real-world applications. Meanwhile, we observed the trend in runtime when using main station clustering on problems with 100-200 stations and it scaled in similar trend with respect to using v.s. not using main stations.}
\begin{figure*}[!h]   
  \begin{minipage}[b]{0.49\textwidth} % 如果一行放2个\UTF{56FE}，用0.5，如果3个\UTF{56FE}，用0.33  
    \centering   
    \includegraphics[width=.9\textwidth]{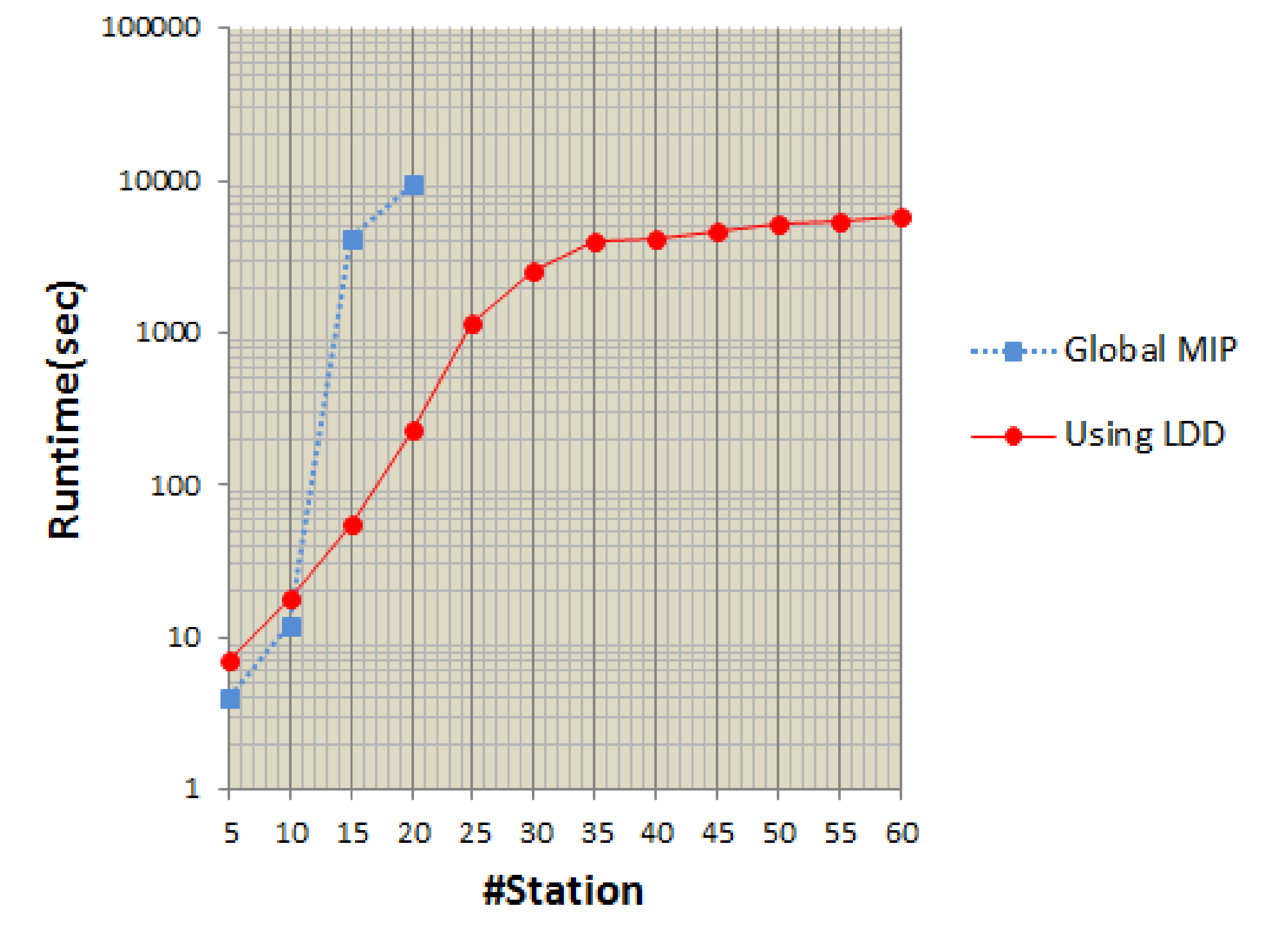}  
    \centerline{(a)}  
  \end{minipage}% 
   \hfill  %水平填充 
  \begin{minipage}[b]{0.49\textwidth}   
    \centering   
    \includegraphics[width=.9\textwidth]{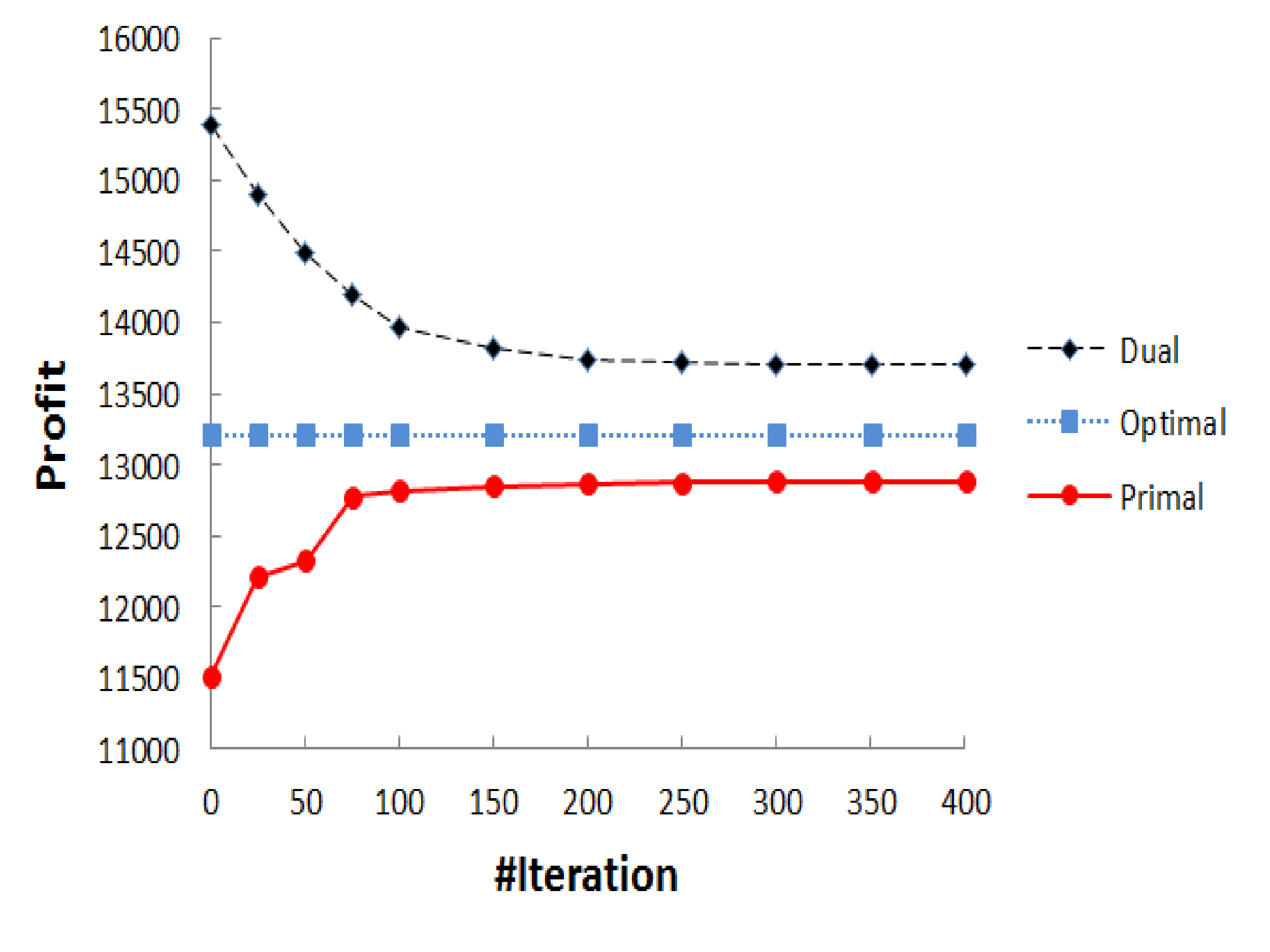}   
    \centerline{(b)}  
  \end{minipage}  
  \caption{(a) Runtime comparison between MILP and LDD, (b) duality gap in the synthetic dataset with 30 stations} \label{fig:2}
\end{figure*} 
\\
\textbf{Duality gap:} We demonstrate the convergence of LDD to near optimal solutions. LDD achieves an optimal solution if the duality gap, i.e., the gap between primal and dual solutions, becomes zero. Figure (\ref{fig:2}b) shows that the duality gap for the instances with 30 stations (grouped into 6 main stations). For these larger problems we are able to obtain a solution with the duality gap of less than 1\%.
\\
\textbf{Main stations:} We also would like to demonstrate the performance of the clustering method in comparison with the optimal solution of instances with 30 base stations (grouped into 6 main stations). Table 3 shows the effect of using main stations on the generated profit and runtime based on five random scenarios of customer demand. With main stations, there is obviously an improvement of more than 13\% in profit on average over all of the optimal solutions from Table 3. Since main stations are based on geographical proximity, it is ideally suitable for handling such scenarios.  
}
\ignore{
\ralf{Furthermore, We have observed more short distance repositioning using carrier vehicles. The key reason behind this negligible loss of demand (lower profit) when using the main stations is the specific demand patterns observed in the real-world data sets. In the actual conditions of the bikeshare networks, the stations that become empty in a particular time period are typically close to each other and hence can be rebalanced within a time step. Since main stations is based on geographical proximity, it is ideally suited to handle such situations. The solution quality of our geographical proximity based on main stations deteriorates if most of the main stations become empty at the same time. To demonstrate this situation, we generate demand scenarios where we read just demand and have high demand for one random station in each main station. Table 3 demonstrates the performance of main stations approach with comparison to the optimal solution. We show that our geographical proximity based on main stations significantly outperforms the existing benchmark approaches due to the specific demand patterns observed in both the real-world data sets of our study. Our solution approaches can be used to group base stations to reduce the size of the {\tt DRRPVT}.}}

\begin{table}
\caption{Effect of main stations with 30 base stations (MS indicates ``main stations'')}\label{30stations}
\centering 
\small
\begin{tabular}{|c|c|c|c|c|c|}
\hline
\multirow{2}*{\thead{Instance}} & \multicolumn{2}{c|}{\thead{With MS}} &  \multicolumn{2}{c|}{\thead{Without MS}} & \multirow{2}*{\thead{profit \\ increase}}   \\ 
\cline{2-5}
         & \thead{Profit} & \thead{Runtime} & \thead{Profit} & \thead{Runtime}& \\
\hline 
1 & 16576 & 37 & 14635 & 1754 & 13.26\% \\
\hline
2 & 16897 & 49 & 14882 & 1774 & 13.54\%\\
\hline
3& 16628 & 41 & 14672 & 1761 & 13.33\%\\
\hline
4 & 16511 & 43 & 14560 & 1762 & 13.40\%\\
\hline
5 & 16134 & 31 & 14212 & 1759 & 13.52\%\\
\hline
\thead{Average} & 16549 & 40 & 14592 & 1762 & 13.41\%\\
\hline
\end{tabular}
\end{table}

\subsection{Varying numbers of stations, vehicles and trailers}
\begin{figure*}[bhtp]   
  \begin{minipage}[b]{0.33\linewidth} % 如果一行放2个\UTF{56FE}，用0.5，如果3个\UTF{56FE}，用0.33  
    \centering   
    \includegraphics[width=0.99\textwidth]{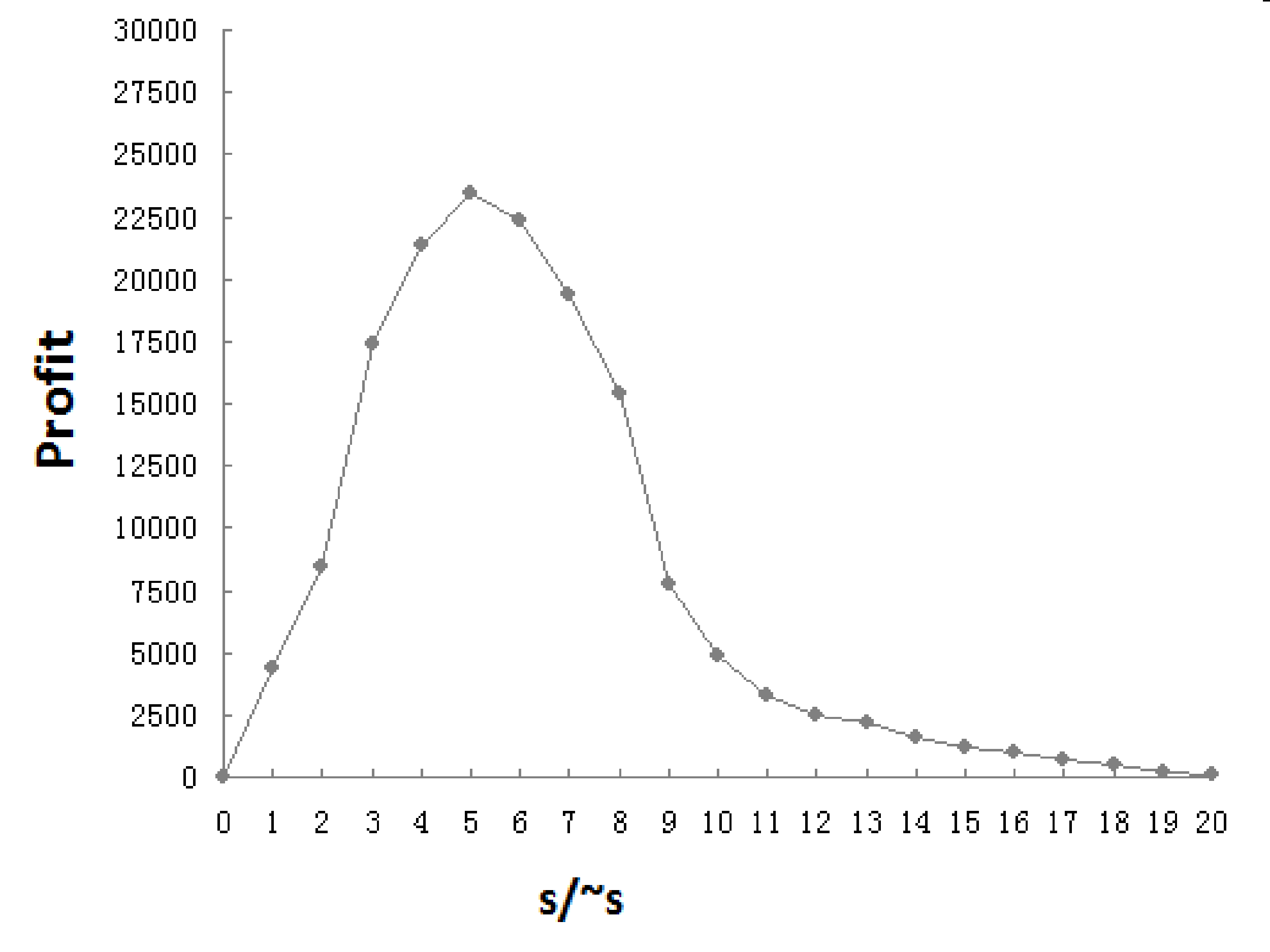}  
    \centerline{(a)}  
  \end{minipage}% 
   \hfill  %水平填充 
  \begin{minipage}[b]{0.33\linewidth}   
    \centering   
    \includegraphics[width=0.99\textwidth]{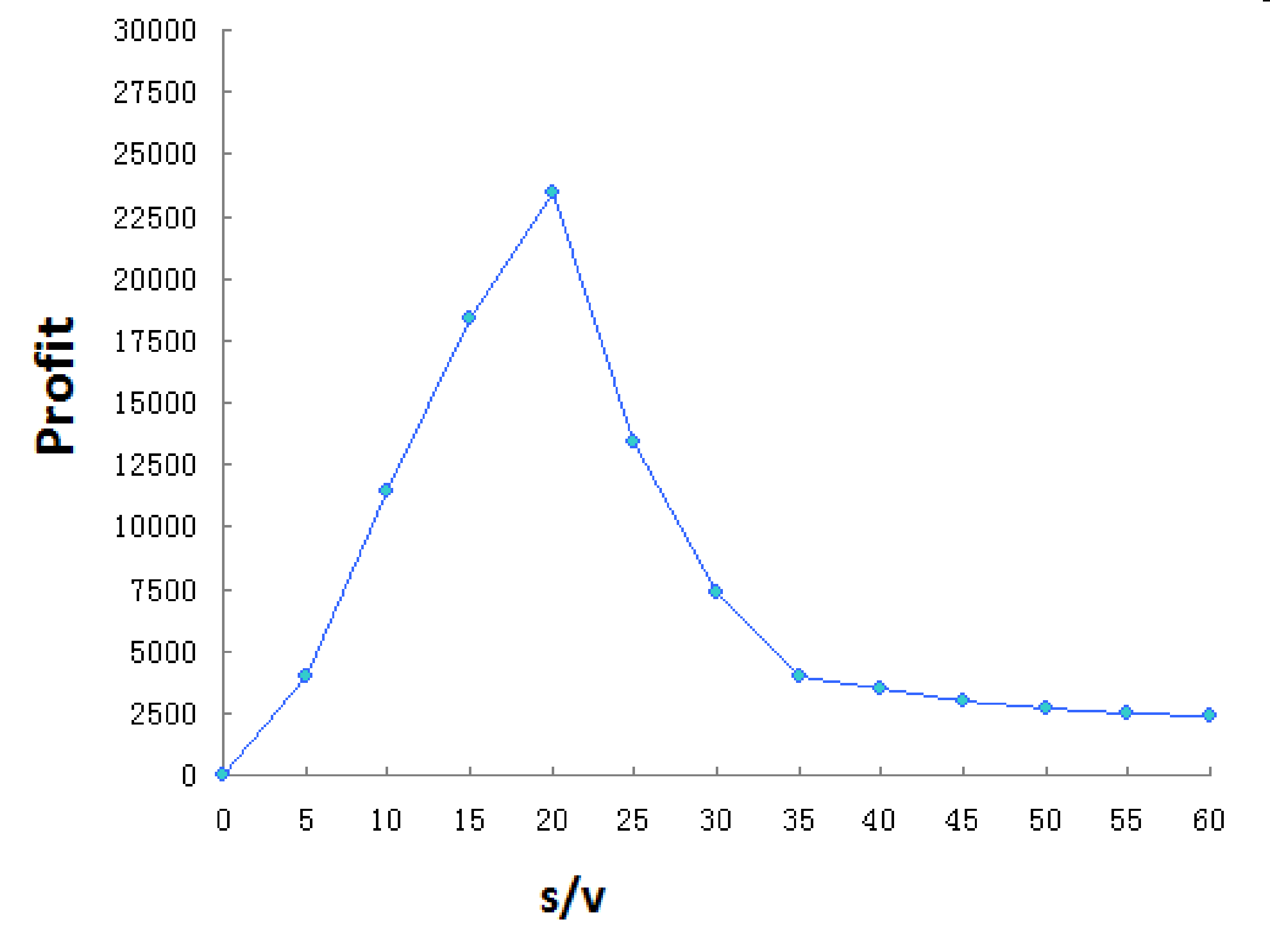}   
    \centerline{(b)}  
  \end{minipage}  
  \hfill  %水平填充 
  \begin{minipage}[b]{0.33\linewidth}   
    \centering   
    \includegraphics[width=0.99\textwidth]{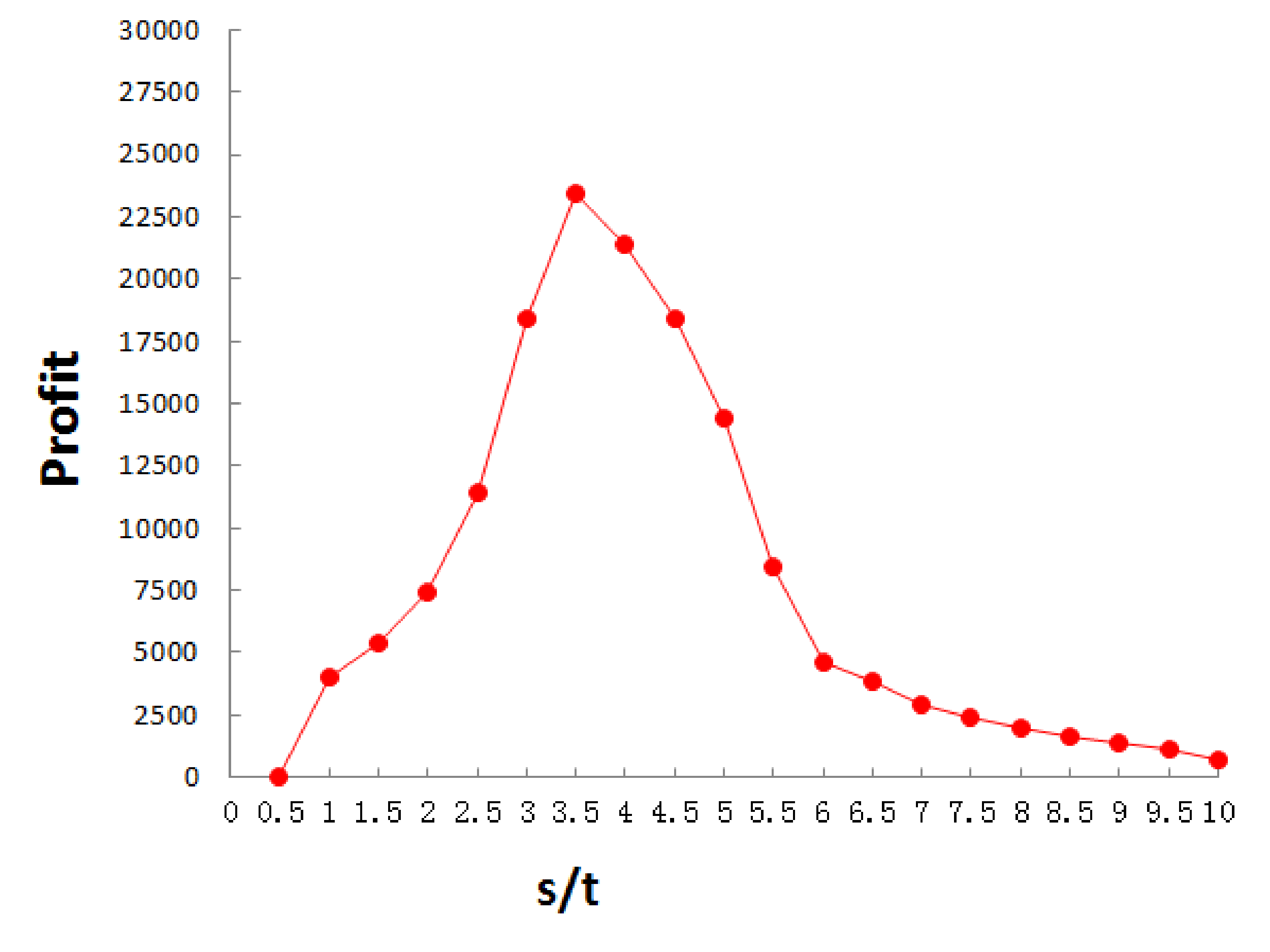}   
    \centerline{(c)}  
  \end{minipage}  
  \caption{The maximal profit of the {\tt DRRPVT} with the ratio of base stations to: (a) main stations, (b) vehicles, and (c) trailers} \label{fig:3}
\end{figure*}

\ignore{The maximum total profit $U_s^{t}$ is achieved if and only if the number of base stations $s$ are five times as likely as the number of main stations $\widetilde{s}$, base stations $s$ are twenty times as likely as the number of carrier vehicles, base stations $s$ are three times as likely as the number of bike trailers. At that point, $U_s^{t}$ achieves the value of maximum.}
\ralf{
We compare the profit of {\tt DRRPVT} with the ratio of base stations to main stations, as shown in Figure (\ref{fig:3}a). The X-axis denotes the ratio of base stations $s$ over main stations $\widetilde{s}$. The Y-axis denotes the total profit. We then compare the profit of the {\tt DRRPVT} with the ratio of base stations over carrier vehicles, as shown in Figure (\ref{fig:3}b). Finally, we evaluate the profit of the {\tt DRRPVT} with the ratio of base stations over bike trailers, as shown in Figure (\ref{fig:3}c). 

From Figure \ref{fig:3}, we can see that the profit of our {\tt DRRPVT} approach generally increases at the beginning, with respect to the increase of the ratios of base stations over main stations, carrier vehicles and bike trailers, respectively. After the profit reaches the maximal value, it goes down when the ratios increase. This is consistent with our intuition since more base stations can indeed raise the profit on repositioning bikes at the beginning. It will, however, largely raise the cost of repositioning bikes when base stations become too many.} 
\ignore{
We obtain: $ U = R \times X - \hat{P}\times f_{v}(\widetilde{s},v)- P \times X \times f_{t}(s,t) $ where $X$ denotes the actual demand of hired bikes belongs to the poisson distribution.$R,\hat{P}, P$ denote the value of hired bikes, trailer tasks and vehicle tasks are constants. $\widetilde{s}$, $s$, $v$, $t$ denote the number of main stations, base stations, vehicles. }
\ignore{
\ralf{Theorem is derived from empirical plotting of the profit metric against various system parameters based on the output from LDD method. Therefore, it is based on theoretical characterization of “the optimal” solution in which one gets maximum benefit.}}

\section{Conclusion}
In this paper we propose an optimization model to jointly consider the usage of carrier vehicles and bike trailers. We build a profit objective to calculate the value of carrier vehicle routing and bike trailers by considering a variety of constraints with respect to vehicle routing and bike repositioning. In the experiment, we exhibit that our approach is effective with comparison to baselines. In the future, it would be interesting to study a budget feasible mechanism which solves the uncertainties in completion time of trailer tasks \ralf{and build an iterative scenario generation approach which provides the update strategies for pre-planned solutions.} In this work, we consider building an objective function and optimizing the objective according to a set of constraints. The constraints are numerous and sometime difficult to create by hand. It would be interesting to study the feasibility of exploiting classical planning models, such as PDDL \cite{DBLP:journals/jair/Geffner03}, with state-of-the-art PDDL model learning approaches \cite{DBLP:journals/ai/ZhuoYHL10,DBLP:conf/ijcai/ZhuoNK13,DBLP:conf/ijcai/ZhuoK13,DBLP:journals/ai/Zhuo014,DBLP:conf/aaai/Zhuo15,DBLP:journals/ai/ZhuoK17} to learn PDDL models from training data automatically, instead of building constraints manually.

\ignore{
%which solves that an adversary identifies the worse demand scenario for a given repositioning strategy. 
\subsubsection{Acknowledgments.}
We would like to express our sincere appreciation to the anonymous reviewers for their insightful comments, which have greatly aided us in improving the quality of the paper. 
}
%%%%%%%%%%%%%%%%%%%%%%%%%%%%%%%%%%%%%%%%%%%%%%%%%%%%%%%%%%%%%%%%%%%%%%%%%%%%%%%%%%%%%
%\subsubsection{The Following Must Conclude Your Document}
%\begin{quote}
%\begin{scriptsize}\begin{verbatim}
% References and End of Paper
% These lines must be placed at the end of your paper
%\bibliography{Bibliography-File}
%\bibliographystyle{aaai}
%\end{document}
%\end{verbatim}\end{scriptsize}
%\end{quote}

%\newpage
\bibliographystyle{aaai}
\bibliography{aaai20}

\end{document}